\documentclass{llncs}
\usepackage{llncsdoc}
\usepackage{subfigure}
\usepackage{cite}
\usepackage{graphicx}
\usepackage{textcomp}
\usepackage{subfigure}
\usepackage{blindtext}
\usepackage{algorithm,algorithmic}
\usepackage[namelimits]{amsmath} 
\usepackage{amssymb}             
\usepackage{amsfonts}            
\usepackage{mathrsfs}            
\usepackage[justification=centering]{caption}
\begin{document}
\markboth{\LaTeXe{} Class for Lecture Notes in Computer
Science}{\LaTeXe{} Class for Lecture Notes in Computer Science}
\thispagestyle{empty}

\vfill
\newpage

\title{VoxelAtlasGAN: 3D Left Ventricle Segmentation on Echocardiography with Atlas Guided Generation and Voxel-to-voxel Discrimination}
\author{Suyu Dong\inst{1} ,\, Gongning Luo\inst{1} ,\,Kuanquan Wang\inst{1} ,\,Shaodong Cao\inst{2} ,\,Ashley Mercado\inst{3,4} ,\, Olga Shmuilovich\inst{3,4} ,\, Henggui Zhang*\inst{1,5},\,Shuo Li\inst{3,4} }

\institute{1. Harbin Institute of Technology, Harbin, China\\
2.Department of Radiology, The Fourth Hospital of Harbin Medical University\\
3.Department of Medical Imaging, Western University, London, Ontario, Canada\\
4.Digital Image Group (DIG), London, ON N6A 3K7, Canada\\
5.University of Manchester, Manchester, UK\\
The corresponding email:henggui.zhang@manchester.ac.uk
}

\maketitle
\begin{abstract}
  3D left ventricle (LV) segmentation on echocardiography is very important for diagnosis and treatment of cardiac disease. It is not only because of that echocardiography is a real-time imaging technology and widespread in clinical application, but also because of that LV segmentation on 3D echocardiography can provide more full volume information of heart than LV segmentation on 2D echocardiography. However, 3D LV segmentation on echocardiography is still an open and challenging task owing to the lower contrast, higher noise and data dimensionality, limited annotation of 3D echocardiography. In this paper, we proposed a novel real-time framework, i.e., VoxelAtlasGAN, for 3D LV segmentation on 3D echocardiography. This framework has three contributions: 1) It is based on voxel-to-voxel conditional generative adversarial nets (cGAN). For the first time, cGAN is used for 3D LV segmentation on echocardiography. And cGAN advantageously fuses substantial 3D spatial context information from 3D echocardiography by self-learning structured loss; 2) For the first time, it embeds the atlas into an end-to-end optimization framework, which uses 3D LV atlas as a powerful prior knowledge to improve the inference speed, address the lower contrast and the limited annotation problems of 3D echocardiography; 3) It combines traditional discrimination loss and the new proposed consistent constraint, which further improves the generalization of the proposed framework. VoxelAtlasGAN was validated on 60 subjects on 3D echocardiography and it achieved satisfactory segmentation results and high inference speed. The mean surface distance is 1.85 mm, the mean hausdorff surface distance is 7.26 mm, mean dice is 0.953, the correlation of EF is 0.918, and the mean inference speed is 0.1s. These results have demonstrated that our proposed method has great potential for clinical application.

\end{abstract}

\section{Introduction}
3D left ventricle segmentation on echocardiography, which directly uses full volume as input, is very important for diagnosis and treatment of cardiac disease. This is because of echocardiography now is the most widely used imaging modality by clinician\cite{pedrosa2016cardiac}, and 3D LV segmentation on echocardiography has an inherent advantage on 3D spatial context information, which is capable of providing more anatomical structure information for clinician. For example, it is possible that there is no LV boundary in some 2D echocardiography slices. While 3D echocardiography combines the full context information, so we can infer the disappeared boundary based on context information. Hence, 3D left ventricle segmentation on echocardiography, compared to the traditional 2D LV segmentation on echocardiography, has more clinical value and has become a hot topic nowadays.

However, 3D LV segmentation on echocardiography is still an open and challenging task owing to the following intrinsic limitations: lower contrast, higher noise and data dimensionality, limited annotation of 3D echocardiography. 1) 3D echocardiography has a lower contrast between the LV borders and surrounding tissue than 2D echocardiography \cite{lang2012eae}, so that 3D LV segmentation will be more likely to have leakage and shrinkage. 2) 3D echocardiography has the higher dimensionality and is more complex than 2D echocardiography, so obtaining better expressive features to achieve accurate segmentation with high processing speed is difficult. 3) Up to now, annotated 3D echocardiography are limited, so it is difficult to train a model to achieve high generalization on 3D LV segmentation task based on the limited annotation 3D echocardiography. Hence, how to utilize 3D echocardiography's advantages and overcome its difficulties to get more accurate 3D LV segmentation results is an urgent problem.
 
Although some methods have been proposed to segment 3D LV on echocardiography\cite{leung2010automated}, these methods are still difficult to overcome the above challenging completely \cite{dong2016left}. Deformable models and statistical models are widely used for the LV segmentation in echocardiography \cite{barbosa2013fast}. However, these methods are sensitive to the large variations of intensity and unable to deal with the low contrast images. Machine learning methods, which use handcrafted features usually, also have been used to segment 3D LV\cite{yang2011prediction}, yet they have limited representation capability to represent the higher dimensionality 3D echocardiography data, complex structure and appearance variations between different subjects. Recently, deep convolutional neural networks (CNNs) \cite{luo2017multi} have been implemented for 3D LV segmentation and it improve the accuracy\cite{oktay2018anatomically}, yet they need large computation depending on the complex 3D CNNs's structure.

In this paper, we proposed a novel automated framework (VoxelAtlasGAN) for 3D LV segmentation on echocardiography. The proposed framework has three contributions: \textbf{1) It uses conditional generative adversarial nets (cGAN) \cite{mirza2014conditional} on voxel-to-voxel mode, which is used for 3D LV segmentation on echocardiography for the first time. Hence, the proposed framework advantageously fuses substantial 3D spatial context information from 3D echocardiography by self-learning structured loss; 2) For the first time, based on cGAN, it embeds the atlas segmentation problem into an end-to-end optimization framework and which uses 3D LV atlas as powerful prior knowledge. In this way, it not only improves the inference speed but also addresses the lower contrast and the limited annotation problems of 3D echocardiography; 3) It combines the traditional discrimination loss and new proposed consistent constraint, which further improves the generalization of the proposed framework.}  Experiment results have demonstrated that our proposed method can obtain satisfactory segmentation accuracy and has the potential for clinical application.

\section{Methods}
\begin{figure}
\vspace{-0.8cm}
\setlength{\abovecaptionskip}{0.2cm} 
\setlength{\belowcaptionskip}{-0.5cm}
\centering
\includegraphics[height=6.5cm,width=11.5cm]{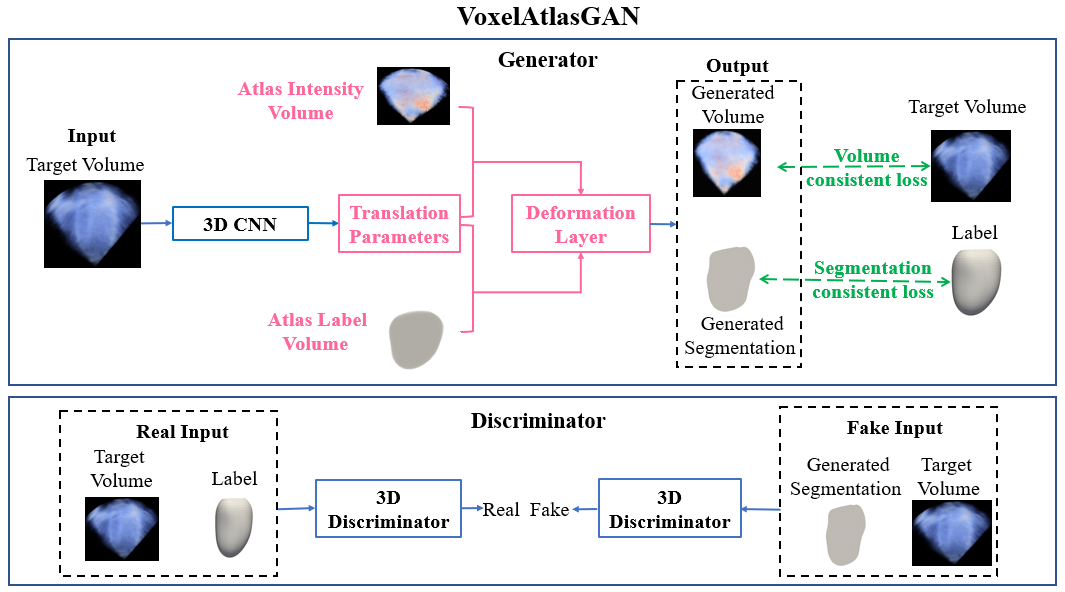}
\caption[]{Framework of the proposed VoxelAtlasGAN.} 
\end{figure}

VoxelAtlasGAN is demonstrated in Fig. 1. Different from usual cGAN, VoxelAtlasGAN adopts a voxel-to-voxel cGAN for high-quality 3D LV segmentation on echocardiography. It uses atlas to provide prior knowledge for 3D generator to guide segmentation, and it combines consistent constraint with discrimination loss as the final optimization object to improve the generalization.
\subsection{Voxel-to-voxel cGAN for high-quality 3D LV segmentation}
Voxel-to-voxel cGAN is designed to automatically achieve high-quality 3D LV segmentation on echocardiography. It adopts full volume as the input which is useful to acquire and exploit substantial 3D spatial context information in 3D echocardiography, hence we call it `voxel-to-voxel' cGAN.

\textbf{Advantages of cGAN compared with CNNs}: 1) The recently proposed cGAN can automatically learn loss function which satisfyingly adapts to data. For segmentation problem, cGAN learns a loss that tries to discriminate the output which is ground truth label or segmentation result from generator, and this loss is minimized by training a generative model. 2) cGAN learns a structured loss, which considers each output pixel depending on the other pixels from the input image, to penalize any possible structure differences between output and target. Hence, the proposed method is based on cGAN for high-quality 3D LV segmentation.
 
\textbf{Structure of VoxelAtlasGAN}: Based on the above advantages of cGAN, we designed novel networks for VoxelAtlasGAN. To address the lower contrast and the limited annotation problems of 3D echocardiography, the proposed method combines atlas prior knowledge, i.e., we formulate the traditional atlas segmentation procedure into the proposed end-to-end deep learning framework (which will be detailed in section 2.2). The specific structure of VoxelAtlasGAN is following.
\textbf{ Generator:} it has five 3D convolution layers, one fully connected layer, and one atlas deformation layer. Specially, the fully connected layer has $w$ nodes, and the $w$ is the number of parameters for translation function of atlas. Besides, the atlas deformation layer models the deformation registration procedure of atlas segmentation. And 3D echocardiography volumes are directly used as the input of the generator.
\textbf{Discriminator:} the proposed discriminator also has five 3D convolution layers and one fully connected layer with two nodes (to classify real or fake). And we use 3D echocardiography volume with the corresponding label and 3D echocardiography volume with the corresponding segmentation result as the input of discriminator. 
\textbf{The optimization object:} the proposed VoxelAtlasGAN combines discrimination loss and the new proposed consistent constraint (consistent segmentation constraint and consistent intensity volume constraint) as the optimization object, which will be detailed in section 2.3.  

\subsection{Atlas guided generation}

\textbf{Advantages of atlas guided generation:} Atlas prior knowledge is able to provide basic anatomical shape information of LV, and incorporating such prior knowledge is crucial for the improvement of LV segmentation performance. Hence, the proposed method embeds the atlas into end-to-end optimization of cGAN framework to provide powerful prior knowledge to overcome some problems. 1) Embedding the atlas prior into cGAN addresses the lower contrast problem of 3D echocardiography through direct registration segmentation method. 2) Embedding the atlas prior into cGAN addresses the limited annotation data problem, due to the proposed method no longer depends on the large numbers of training data to model the shape prior. 3) Embedding the atlas prior into cGAN avoids the complex and time-consuming computation of traditional atlas-based methods, because of that the proposed method formulates the transformation parameters optimization problem into the end-to-end deep learning framework (cGAN), which is based on the differentiability of transformation parameters on atlas segmentation procedure. 4) Embedding the atlas segmentation procedure into cGAN enhances the interpretability of cGAN.

\textbf{Atlas deformation layer:} Specifically, the transformation function of atlas segmentation includes global rigid transformation $T_{global}$ using affine transformation and local non-rigid transformation $T_{local}$ using free-form deformations (FFD) method \cite{rueckert1999nonrigid}. Hence, we model the atlas deformation layer as following:
\begin{equation}
T(X)=T_{local}(T_{global}(X)), X\in \left \{ Atlas_{label\_volume}, Atlas_{intensity\_volume} \right \}
\end{equation}
\begin{equation}
T_{global}(X)=\begin{bmatrix}
\theta_1, & \theta_2, & \theta_3\\ 
 \theta_4, &  \theta_5, & \theta_6 \\ 
 \theta_7, & \theta_8,  & \theta_9
\end{bmatrix}*X+\begin{bmatrix}
\theta_{10}\\ 
\theta_{11}\\ 
\theta_{12}
\end{bmatrix}
\end{equation}
\begin{equation}
T_{local}(T_{global}(X))=B_\phi(T_{global}(X))
\end{equation}
where $T$ denotes the final deformation function for atlas transformation, $X$ denotes the 3D LV atlas including atlas label volume $Atlas_{label\_volume}$ and atlas intensity volume $Atlas_{intensity\_volume}$, $\theta$ is rigid affine transformation parameter set including 12 parameters, $B$ is the B-spline function, and $\phi$ is the control point set (which is the parameter set of non-rigid FFD transformation). Specially, B-spline FFD is initialized by 1000 equidistant control points with 3000 parameters to control deformation. To embed the 3012 transformation parameters into end-to-end cGAN, we make the output of the fully connected layer(which in the generator network of cGAN) has 3012 nodes. Hence, 3012 deformation parameters of transformation function are formulated as corresponding 3012 latent variables, which are solved by unified deep learning optimization framework. Besides, in this way, model's interpretability is strong compared with common deep learning methods, because the 3012 deformation parameters are inherently clear for transformation function for atlas deformation layer.

Additionally, in this work, atlas was built from the mean space of a set of 3D echocardiography. To avoid segmentation biases to specific noise and artifacts, we employed 3D echocardiographies which were acquired using two devices to construct atlas. They were registered to mean space. Atlas intensity volume and atlas label volume were computed from this set of registered 3D echocardiography and corresponding registered segmentation results respectively. In the test stage, based on atlas and the learned translation parameters, we obtained the final 3D LV segmentation results through the generator of VoxelAtlasGAN. 

\subsection{ Voxel-to-voxel discrimination with consistent constraint}
Based on the shared deformation layer, the consistent constraint, including segmentation consistent constraint and volume consistent constraint, is proposed as a part of cGAN's loss function to naturally guarantee the generalization of the proposed framework and further improve the segmentation accuracy. The segmentation consistent constraint measures the similarity between the ground truth labels and the generated segmentation results. Analogously, the volume consistent constraint measures the similarity between the input volumes and the generated intensity volumes. Hence, the final optimization object of the proposed VoxelAtlasGAN is :
\begin{equation}
\arg \min_G \max_DL_{VoxelAtlasGAN}= L_{cGAN}\left ( G,D \right )+\alpha L_{label}\left ( G \right )+\beta L_{intensity}\left ( G \right )
\end{equation}
where $G$ denotes the generator of VoxelAtlasGAN, $D$ denotes the discriminator of VoxelAtlasGAN, $L_{cGAN}\left ( G,D \right )$ denotes the cGAN loss, $L_{label}\left ( G \right )$ and $L_{intensity}\left ( G \right )$ denote the segmentation consistent constraint and volume consistent constraint with weights $\alpha$ and $\beta$ respectively.
The cGAN loss is: 
\begin{equation}
L_{cGan}(G,D)= \log D\left ( x,y \right )+\log \left ( 1-D\left ( x,G_{label\_volume} \right ) \right )
\end{equation}
where $G$ and $D$ have the same denotations with above equations (4), $x$ denotes the target volume, $y$ denotes the ground truth label, and $G_{label\_volume}$ is the generated segmentation result from generator.
Specially, segmentation consistent constraint is modeled by L1 norm:
\begin{equation}
L_{label}\left ( G \right )= \left \| y-G_{label\_volume} \right \|_1
\end{equation}
Because the 3D echocardiography exist large random noisy and high intensity variability, the volume consistent constraint is modeled through normalized mutual information method \cite{zhuang2010registration}, which has high robustness on similarity measurement. The volume consistent constraint is:
\begin{equation}
L_{intensity}\left ( G \right )=-(H(x)+H(G_{intensity\_volume}))/H(x,G_{intensity\_volume})
\end{equation}
where $G_{intensity\_volume}$ denotes the intensity volume from generator, $H(x)$ and $H(G_{intensity\_volume})$ denote the marginal entropies of $x$ and $G_{intensity\_volume}$ respectively, and $H(x,G_{intensity\_volume})$ denotes the joint entropies of $x$ and $G_{intensity\_volume}$. 

\section{Dataset and Setting}
VoxelAtlasGAN is validated on 3D echocardiography with 25 training subjects and 35 validation subjects. Each subject includes two labeled volumes in the end-systole (ES) and end-diastole (ED) frames. The labels are obtained by three clinicians and the final labels are mutually authenticated.
We adopt the same cGAN training mode with \cite{mirza2014conditional}. We use SGD solver, the learning rate is 0.0002, the momentum is 0.5, and the batch size is 1. At inference time, we run the generator net in the same manner as the training phase. The weights $\alpha$ and $\beta$ for segmentation consistent constraint and volume consistent constraint are 0.6 and 0.4 respectively. The proposed VoxelAtlasGAN was implemented based on the widely used pytorch framework and the whole experiment was performed on NVIDIA Titan X GPU.

We adopt the evaluation criterions in \cite{bernard2016standardized} to evaluate our proposed method, which include mean surface distance(MSD), mean hausdorff surface distance(HSD), mean dice index(D), and correlation (corr) of ejection fractions (EF). 
\section{Results and analysis}
VoxelAtlasGAN's segmentation performance is powerfully supported in Table. 1. 1) In the aspect of segmentation accuracy, we find that VoxelAtlasGAN achieves the best segmentation results. The mean surface distance, hausdorff surface distance and dice are 1.85 mm, 7.26 mm and 0.953 respectively. And the correlation of EF (between the segmentation results and the ground truth ) is 0.918 and the corresponding standard deviation is 0.016. These results prove the superiority of the proposed VoxelAtlasGAN compared with the existing methods. 2) In the aspect of segmentation efficiency, the proposed method achieves the best segmentation speed, it only needs 0.1s for segmentation of every volume, which is crucial for real-time clinical application. 

We also evaluated the importance of atlas prior and the proposed consistent constraint by ablation experiments (in the last three rows of Table. 1), which replace or remove a single component from our framework. The results showed that atlas prior improved the segmentation accuracy apparently, and the proposed consistent constraint also brought important improvement for segmentation. What's more, compared to the 3D CNN method, which adopts 3D full convolution network(V-net\cite{milletari2016v}) to segment 3D LV on echocardiography, the proposed voxel-to-voxel cGAN is superior even without the atlas and the proposed consistent constraint. 
\begin{table}
\vspace{-0.8cm}
\setlength{\abovecaptionskip}{0.2cm} 
\setlength{\belowcaptionskip}{0.6cm}
\caption{The segmentation performance comparison among existing methods and the proposed VoxelAtlasGAN under different configurations.(3D cGAN denotes the VoxelAtlasGAN without atlas prior and consistent constraint, VoxelGANWA denotes VoxelAtlasGAN without consistent constraint.)}
\begin{center}
\renewcommand{\arraystretch}{1.4}
\setlength\tabcolsep{3pt}
\begin{tabular}{lp{2cm}lp{2cm}lp{2cm}lp{2cm}lp{2cm}lp{2cm}l}
\noalign{\smallskip}
\hline
\noalign{\smallskip}
  Methods & D & MSD(mm) & HSD(mm) & Corr of EF & speed (s)\\
 \hline
 3D Atlas\cite{oktay2014learning}  & 0.88$\pm$0.03 & 2.26$\pm $0.74 & 9.92$\pm $2.16 & 0.836$\pm $0.079 & 2000\\
 V-net\cite{milletari2016v} & 0.89$\pm$0.035 & 2.1$\pm $0.71  & 9.79$\pm $8.9 & 0.86$\pm $0.053   & 20\\
 
3D cGAN & 0.914$\pm$0.028 & 1.98$\pm $0.65  & 8.91$\pm $7.3 & 0.89$\pm $0.032 &10\\
 
VoxelGANWA & 0.939$\pm$0.021 & 1.93$\pm $0.52 & 8.35$\pm $4.57 & 0.907$\pm $0.029 & 0.1\\

VoxelAtlasGAN & \textbf{0.953$\pm$0.019} & \textbf{1.85$\pm $0.43}  & \textbf{7.26$\pm $2.3} & \textbf{0.918$\pm $0.016} & \textbf{0.1}\\
\hline
\end{tabular}
\end{center}
\vspace{-1.2cm}
\end{table}
\section {Conclusions}
In this paper, we proposed a novel automated framework VoxelAtlasGAN for high-quality 3D LV segmentation on echocardiography. The proposed framework used voxel-to-voxel cGAN to advantageously fuse substantial 3D spatial context information from 3D echocardiography for the first time. Besides, it embedded the powerful atlas prior knowledge into end-to-end optimization framework for the first time. It also combined the new proposed consistent constraint and traditional discrimination loss as the final optimization object to further improve the generalization of the proposed framework. Experiment results have demonstrated that our proposed method obtained satisfactory segmentation accuracy and has potential of clinical application.

\section*{Acknowledgments}

This work was supported by the Natural Science Foundation of China under Grant No. 61572152 and 61571165.

\bibliographystyle{splncs}
\bibliography{reference}

\begin{thebibliography}{10}

\bibitem{pedrosa2016cardiac}
Pedrosa, J., Barbosa, D., et~al.:
\newblock Cardiac chamber volumetric assessment using 3d ultrasound-a review.
\newblock Current pharmaceutical design \textbf{22}(1) (2016)  105--121

\bibitem{lang2012eae}
Lang, R.M., Badano, L.P., et~al.:
\newblock Eae/ase recommendations for image acquisition and display using
  three-dimensional echocardiography.
\newblock Journal of the American Society of Echocardiography \textbf{25}(1)
  (2012)  3--46

\bibitem{leung2010automated}
Leung, K.E., Bosch, J.G.:
\newblock Automated border detection in three-dimensional echocardiography:
  principles and promises.
\newblock European journal of echocardiography \textbf{11}(2) (2010)  97--108

\bibitem{dong2016left}
Dong, S., Luo, G., Sun, G., Wang, K., Zhang, H.:
\newblock A left ventricular segmentation method on 3d echocardiography using
  deep learning and snake.
\newblock In: Computing in Cardiology Conference (CinC), 2016, IEEE (2016)
  473--476

\bibitem{barbosa2013fast}
Barbosa, D., Dietenbeck, T., et~al.:
\newblock Fast and fully automatic 3-d echocardiographic segmentation using
  b-spline explicit active surfaces: feasibility study and validation in a
  clinical setting.
\newblock Ultrasound in Medicine and Biology \textbf{39}(1) (2013)  89--101

\bibitem{yang2011prediction}
Yang, L., Georgescu, B., et~al.:
\newblock Prediction based collaborative trackers (pct): A robust and accurate
  approach toward 3d medical object tracking.
\newblock IEEE Transactions on Medical Imaging \textbf{30}(11) (2011)
  1921--1932

\bibitem{luo2017multi}
Luo, G., Dong, S., Wang, K., Zuo, W., Cao, S., Zhang, H.:
\newblock Multi-views fusion cnn for left ventricular volumes estimation on
  cardiac mr images.
\newblock IEEE Transactions on Biomedical Engineering (2017)

\bibitem{oktay2018anatomically}
Oktay, O., Ferrante, E., Kamnitsas, K., Heinrich, M., Bai, W., Caballero, J.,
  Cook, S.A., de~Marvao, A., Dawes, T., O‘Regan, D.P.,  et~al.:
\newblock Anatomically constrained neural networks (acnns): Application to
  cardiac image enhancement and segmentation.
\newblock IEEE transactions on medical imaging \textbf{37}(2) (2018)  384--395

\bibitem{mirza2014conditional}
Mirza, M., Osindero, S.:
\newblock Conditional generative adversarial nets.
\newblock arXiv preprint arXiv:1411.1784 (2014)

\bibitem{rueckert1999nonrigid}
Rueckert, D., Sonoda, L.I., Hayes, C., Hill, D.L., Leach, M.O., Hawkes, D.J.:
\newblock Nonrigid registration using free-form deformations: application to
  breast mr images.
\newblock IEEE transactions on medical imaging \textbf{18}(8) (1999)  712--721

\bibitem{zhuang2010registration}
Zhuang, X., Rhode, K.S., Razavi, R.S., Hawkes, D.J., Ourselin, S.:
\newblock A registration-based propagation framework for automatic whole heart
  segmentation of cardiac mri.
\newblock IEEE transactions on medical imaging \textbf{29}(9) (2010)
  1612--1625

\bibitem{bernard2016standardized}
Bernard, O., Bosch, J.G., Heyde, B., Alessandrini, M., Barbosa, D.,
  Camarasu-Pop, S., Cervenansky, F., Valette, S., Mirea, O., Bernier, M.,
  et~al.:
\newblock Standardized evaluation system for left ventricular segmentation
  algorithms in 3d echocardiography.
\newblock IEEE transactions on medical imaging \textbf{35}(4) (2016)  967--977

\bibitem{milletari2016v}
Milletari, F., Navab, N., Ahmadi, S.A.:
\newblock V-net: Fully convolutional neural networks for volumetric medical
  image segmentation.
\newblock In: 3D Vision (3DV), 2016 Fourth International Conference on, IEEE
  (2016)  565--571

\bibitem{oktay2014learning}
Oktay, O., Shi, W., Keraudren, K., Caballero, J., Rueckert, D., Hajnal, J.:
\newblock Learning shape representations for multi-atlas endocardium
  segmentation in 3d echo images.
\newblock Proceedings MICCAI Challenge on Echocardiographic Three-Dimensional
  Ultrasound Segmentation (CETUS), Boston, MIDAS Journal (2014)  57--64

\end{thebibliography}

\end{document}